# The Artificial Intelligence Ontology: LLM-assisted construction of AI concept hierarchies

Marcin P. Joachimiak[1], Mark A. Miller[1], J. Harry Caufield[1], Ryan Ly[2], Nomi L. Harris[1], Andrew Tritt[3], Christopher J. Mungall[1], Kristofer E. Bouchard[2,4,5,6]

[1]Biosystems Data Science Department, Environmental Genomics and Systems Biology Division, Lawrence Berkeley National Laboratory, 1 Cyclotron Road, Berkeley, CA 94720, USA

[2]Scientific Data Division, Lawrence Berkeley National Laboratory, 1 Cyclotron Road, Berkeley, CA 94720, USA

[3]Applied Mathematics and Computational Research Division, Lawrence Berkeley National Laboratory, 1 Cyclotron Road, Berkeley, CA 94720, USA

[4]Biological Systems & Engineering Division, Lawrence Berkeley National Laboratory, 1 Cyclotron Road, Berkeley, CA 94720, USA

[5]Helen Wills Neuroscience Institute, UC Berkeley, Berkeley, CA 94720, USA

[6]Redwood Center for Theoretical Neuroscience, UC Berkeley, Berkeley, CA 94720, USA

## Abstract

The Artificial Intelligence Ontology (AIO) is a systematization of artificial intelligence (AI) concepts, methodologies, and their interrelations. Developed via manual curation, with the additional assistance of large language models (LLMs), AIO aims to address the rapidly evolving landscape of AI by providing a comprehensive framework that encompasses both technical and ethical aspects of AI technologies. The primary audience for AIO includes AI researchers, developers, and educators seeking standardized terminology and concepts within the AI domain. The ontology is structured around six top-level branches: Networks, Layers, Functions, LLMs, Preprocessing, and Bias, each designed to support the modular composition of AI methods and facilitate a deeper understanding of deep learning architectures and ethical considerations in AI.

AIO's development utilized the Ontology Development Kit (ODK) for its creation and maintenance, with its content being dynamically updated through AI-driven curation support. This approach not only ensures the ontology's relevance amidst the fast-paced advancements in AI but also significantly enhances its utility for researchers, developers, and educators by simplifying the integration of new AI concepts and methodologies.

The ontology's utility is demonstrated through the annotation of AI methods data in a catalog of AI research publications and the integration into the BioPortal ontology resource, highlighting its potential for cross-disciplinary research. The AIO ontology is open source and is available on GitHub (https://github.com/berkeleybop/artificial-intelligence-ontology) and BioPortal (https://bioportal.bioontology.org/ontologies/AIO).

## Introduction

The field of Artificial Intelligence (AI) is having a transformational effect on many research domains and human life in general. AI as a discipline encompasses both the computational approaches and social and ethical aspects. Modeling knowledge about AI provides important benefits through standardizing terminology, concepts, and their relationships. Previous work has focused on higher level modeling of machine learning (ML) concepts or more detailed modeling of statistical concepts and methods. As one prominent example, the Machine Learning Ontology (MLOnto) defines seven top-level classes covering both the processes and tools of ML as well as vocabulary and categorizations particular to the field (e.g., a class *MLTypes* defines categories such as *Unsupervised Learning*) (Braga et al., 2020). Blagec et al. pursued similar goals, in their case yielding an Intelligence Task Ontology and Knowledge Graph (ITO) (Blagec et al., 2022). ITO places additional emphasis on measurements and benchmark results, along with relationships between specific performance metrics such as *Area under curve* and *BLEU score.* Other resources such as the NIST AI glossary, focus more on the terminology in the broader AI ecosystem (Schwartz et al., 2023), rather than hierarchies of related methods for example, and do not include relationships. Cross-domain ontologies such as EDAM (Black et al., 2022), the Computer Science Ontology (CSO), and the Software Ontology (SWO) (Malone et al., 2014) also define many of the concepts relevant to AI, though with a focus on their respective domains rather than on AI applications or impact.

In some ways, the field of AI is self-categorizing. Repositories of AI literature and models already describe many concepts in community-defined ways. When the creators of ITO defined sets of benchmarks and tasks, for example, they began by extracting classifications from the Papers with Code resource (*Papers with Code - The Latest in Machine Learning*, n.d.). Since that time, and particularly with the successful application of large language models (LLMs), researchers have contributed massive quantities of new vocabulary to the field and a growing collection of new resources to open repositories (e.g., models and data in Papers with Code, Hugging Face, code in GitHub). Even the term "artificial intelligence" itself has grown to encompass a widening assortment of methods, use cases, and general philosophies (Bearman et al., 2023). The exact relationships between models or methods and their results or publications often remains unspecified.

We developed the Artificial Intelligence Ontology (AIO) primarily to standardize concepts and relationships integral to AI methods, though also to address the need for more holistic considerations in AI applications. We have defined classes to cover more recent LLM advances and earlier approaches. With an eye toward technical applications, we have aligned the ontology with terminology used in ML platforms such as Tensorflow and PyTorch. We have designed AIO to be rapidly extendable and responsive to new innovations in the field. Crucially, we have also included classes related to the ethical and legal impact of AI methods (Bender et al., 2021; Ning et al., 2023) (for example, we define Bias as a top-level class). Finally, AIO was designed for and built with LLM-assisted content suggestion and curation support, which allows the ontology to more easily keep up to date and scale with advances in the AI fields. Collectively, these design decisions enable AIO to satisfy community needs regarding standardization of AI/ML concepts across its numerous areas of application. The AIO is an open source project (https://github.com/berkeleybop/artificial-intelligence-ontology) that encourages the broader community not only to make use of the ontology and related tools, but also to contribute comments, requests, and additions.

# Methods

## Ontology Creation using the Ontology Development Kit

We used the Ontology Development Kit (ODK) (Matentzoglu et al., 2022) to organize the AIO and to set up a GitHub repository (https://github.com/berkeleybop/artificial-intelligence-ontology) containing source components and workflows to build the ontology. The ODK includes a suite of ontology tools, including the very configurable open source ROBOT library (R. C. Jackson et al., 2019). As a result, each AIO build includes artifacts in a variety of serializations (OBO, OWL, and obojson) and variations to support multiple use cases (e.g., a base version without imported axioms). AIO is built from an ontology ROBOT template (Sup. Info.). ODK also enables straightforward reproducibility as it is distributed as a Docker container; any user may build or modify AIO with the same set of open-source tools used in "official" builds.

We make each release of this ontology available through BioPortal (Whetzel et al., 2011). The BioPortal platform provides straightforward navigation of the AIO hierarchy and accessibility through a central API as well as a web UI. Additionally, the Bioportal API enables automatic generation of predicted lexical mappings between AIO classes and those in other ontologies. The ontology is also available as a SQLite database via the semantic-sql repository, allowing for expressive SQL queries, or programmatic access via the OAK library.

## Curation and ingestion of individual branches

Data sources for individual ontology branches are as follows. The Network branch was developed from various sources including publications (Sarker, 2021), the Asimov Institute Neural Network Zoo (van Veen & Leijnen, 2016), and Wikipedia(Wikipedia contributors, n.d.). Names of layers and functions used in AI method development were sourced from pyTorch (Paszke et al., 2019) and TensorFlow (Abadi, 2016) documentation. We included all Layers and Functions from TensorFlow, but for demonstration purposes we included only Normalization and Pooling Layers from PyTorch. LLM and Preprocessing classes were developed using LLM approaches (see *AI-driven curation support*). AI Biases were sourced from a NIST report on AI bias (Schwartz et al., 2022) as well as from Wikipedia. General machine learning methods were sourced from Wikipedia and textbooks (Brownlee, 2013). References for each ontology class are provided in the AIO data serializations produced by the ODK build.

To assist in automation, ROBOT templates (R. C. Jackson et al., 2019) were created for each of the main ontology branches. This allows each branch of the ontology to be compiled from Google Sheets with content generated using LLM prompts that can be easily authored and contributed to by domain experts. Each template was manually created, and the templates were populated using a mixture of automated and manual methods.

The AIO content structured with ROBOT templates was also amenable to ontology construction using LLMs. We developed the LLM and Preprocessing branches of the ontology using Claude 3 Sonnet (March 2024) (*Introducing the next Generation of Claude*, n.d.) and GPT-4 (OpenAI et al., 2023) (March 2024, additionally using the ROBOT-helper chatbot (Creators Mungall, n.d.))

with designed prompts (Sup. Info) and inputting example ontology data rows from the AIO sheet and requesting extension of this data with additional LLM method terms.

## AI-assisted ontology development

AI can support humans in ontology development including for ontology seeding, extension, maintenance, and updating, as seen with recently emerging tools relying on LLMs (Caufield et al., 2023; Toro et al., 2023). The ROBOT ontology creation template enabled us to leverage LLMs to streamline the process of AIO ontology expansion and refinement. This straightforward tabular ontology format is amenable to interactions with LLMs by providing few-shot learning examples or even allowing to input the entire AIO into prompt text. In addition, the AIO ontology format supports interoperability with LLM-based tools (*Curate-Gpt: LLM-Driven Curation Assist Tool (pre-Alpha)*, n.d.; Harry Caufield et al., 2023; Mungall et al., 2022; Toro et al., 2023). This approach allows us to dynamically incorporate new AI concepts and methodologies into AIO as they emerge. In summary, the AI-assisted ontology development allows editors to work with the tabular ontology format (ROBOT template) faster, to extract data from other sources (e.g. images, free text), all under human supervision.

## Evaluation

We focused on a real-world use case evaluation by annotating the methods data from the Papers with Code resource, which collects papers about AI/ML methods. This was performed using the Ontology Access Kit (OAK) framework (*Ontology-Access-Kit: Ontology Access Kit: A Python Library and Command Line Application for Working with Ontologies*, n.d.), using the AIO ontology and data on 2,194 publications describing AI/ML methods from Papers with Code (https://production-media.paperswithcode.com/about/methods.json.gz, accessed on 3/28/24). We used the OAK *annotate* command to identify exact matches of AIO class names as well as their synonyms across the different field values available in the Papers with Code methods data. These results were summarized per AIO ontology class, with the final data shown in Figure 2.

We built the ontology using the Ontology Development Toolkit (ODK) (Matentzoglu et al., 2022), which provides a standard way of organizing ontology content, as well as standard workflows integrated with GitHub actions for structurally and semantically validating the content of an ontology. We configured the ODK to use the ELK reasoner (Kazakov et al., 2014), which provides fast reasoning over the EL ontology profile for validation of AIO. Note that, like many bio-ontologies, the AIO avoids complex OWL-DL constructs outside the EL profile (Motik et al., n.d.). In addition, different data serializations of the AIO are generated during the ontology build allowing for different representations with different levels of expressiveness and human and machine readability. The integration of the AIO into BioPortal (https://bioportal.bioontology.org/ontologies/AIO) is dependent on the OWL serialization and while BioPortal does not require deep ontology validation, the UI aspects of the resource allow an ontology developer to quickly identify different types of ontology issues. The data-driven evaluation of the AIO by annotating Papers with Code relied on the OAK framework, where we used Open Biomedical Ontology OBO format (*The OBO Flat File Format Guide, Version 1.4*, n.d.) files directly or another ontology serialization available in OAK as a SQLite database. We also used the ROBOT statistics command to generate a report for the AIO ontology; this data was used for Table 1. Thus, the AIO ontology was parsed, loaded, and utilized in four different contexts, serving as an evaluation of practical applications. Finally, we used the MIRO guidelines to help improve both the AIO ontology and this manuscript by inputting both the guidelines and the manuscript text into a GPT-4 prompt and asking which guidelines were met and which ones were not. This process was very efficient and effective, allowing to quickly analyze and improve the text and satisfy all MIRO guidelines (Sup. Info.).

### Availability and License

The AIO is shared as an open resource under the Creative Commons Attribution 4.0 International License (CC BY 4.0), allowing for wide reuse and modification with proper attribution. We use BioPortal (Whetzel et al., 2011) as a means of distributing the ontology (https://bioportal.bioontology.org/ontologies/AIO), with BioPortal automatically pulling the ontology data serializations from GitHub releases (https://github.com/berkeleybop/artificial-intelligence-ontology). Development discussions and feature requests can be made through the AIO GitHub issue tracker (https://github.com/berkeleybop/artificial-intelligence-ontology/issues).

# Results

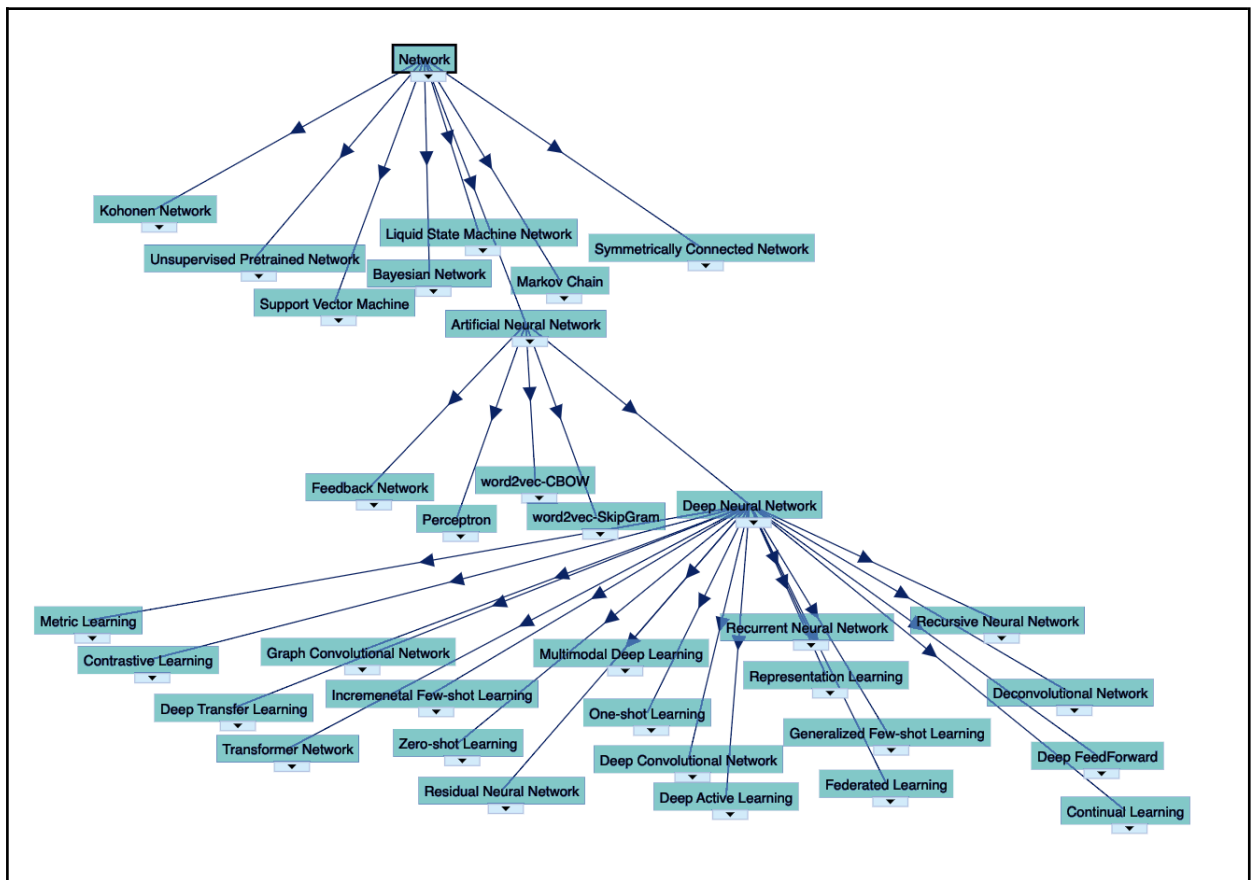

**Figure 1.** Structure of the Network branch of the AIO. For simplicity, not all child nodes are shown.

## High-level ontology structure

The AIO structure consists of six top level branches: Networks, Layers, Functions, LLMs, Preprocessing, and Bias. The Network (Figure 1), Layer, and Function branches are interlinked, with many Network classes having a representation based on a series of Layer terms. Thus, the Layer and Function branches are modeled to support modular composition to enable flexible representations of possible methods based on existing AI development frameworks.We summarize major ontology features for AIO in Table 1. The ontology contains 417 classes, 360 synonyms for these classes, and 414 is_a relationships. AIO outlines the various types of layers (e.g., convolutional, recurrent, pooling) and functions (e.g., activation functions, loss functions) that constitute AI/ML models.

## Representing data content and ethical aspects of AI: bias

Bias in AI research and applications directly impacts the fairness, reliability, and generalizability of AI systems. The standardization of AI concepts, including the identification and mitigation of bias, is crucial for developing AI technologies that are both effective and equitable. The National Institute of Standards and Technology (NIST) report on AI bias provides a taxonomy of bias concepts relevant to AI as well as guidelines for identifying and managing bias in AI systems. The Bias branch in AIO was developed using a NIST report on AI bias (Schwartz et al., 2022) and related Wikipedia entries.

| Table 1. AIO branch summary statistics | | | | | | | | |
|---|---|---|---|---|---|---|---|---|
| Branch summary statistic | Bias | Function | Layer | Large Language Model | Machine Learning | Network | Preprocessing | Total |
| class count | 61 | 14 | 158 | 66 | 51 | 54 | 13 | 417 |
| is_a edge count | 60 | 13 | 157 | 66 | 52 | 54 | 12 | 414 |
| synonym count | 29 | 9 | 117 | 65 | 35 | 82 | 23 | 360 |

AIO supports a more systematic and comprehensive approach to characterizing bias in AI. This branch encompasses various types of biases that can arise throughout the AI development lifecycle, from data collection and model training to the deployment and evaluation of AI systems. It includes the following bias categories: computational, historical, human, institutional, societal, and systemic. Different biases in these categories can significantly affect the performance and fairness of AI applications.

## NLP evaluation

We conducted a Natural Language Processing (NLP) evaluation to assess the coverage and applicability of AIO within the context of practical AI research. This evaluation involved lexical matching of terms from the Papers with Code Methods dataset representing 2,194 research publications reporting AI/ML methods against the term labels and synonyms defined in AIO. The goal was to demonstrate that AIO covers the range of standardized concepts represented in current AI research and development practices as documented in Papers with Code.

The process utilized standard concept recognition techniques to ensure accurate matching, taking into account variations in terminology and the use of synonyms within the AI field (Mungall et al., 2022). This approach allowed us to map the vast array of methods and technologies listed in Papers with Code to the structured hierarchy of AIO, thereby validating the ontology's relevance and utility in categorizing AI methodologies. The Papers with Code methods classification describes seven AI/ML methods areas, and two areas ("Natural Language Processing" and "Reinforcement Learning") are represented in AIO, with "Audio", "Computer Vision", "General", "Graph", "Sequential" not represented in AIO because they are out of scope for AI/ML specific concepts. Further, Papers with Code also has 313 collections and AIO terms represent 37 of these. We note the difference in scope, where Papers with Code are not developing an ontology and represent a controlled vocabulary with categories thus allowing to include terms corresponding to every topic that may be in multiple papers, even automatically. On the other hand, nearly half of AIO terms, 205 out of the 417, were found in paper titles and method classification fields of the Papers with Code data. This represents a high level of coverage of AIO terms given that many Bias, Layer, and Function AIO classes are unlikely to be explicitly represented in this data. These results suggest that while Papers with Code is a good data source for extending the AIO, the AIO can also help inform AI/ML resources about potential other standardized terms and relationships for classification.

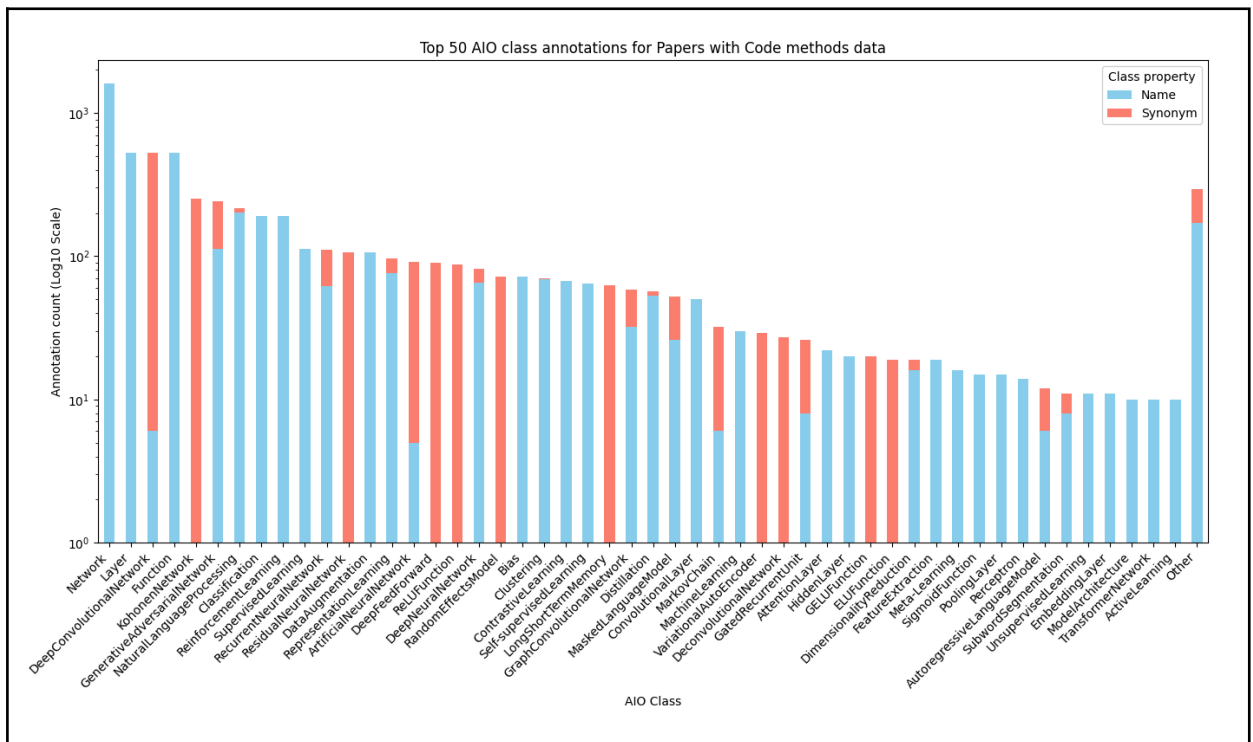

**Figure 2.** The number of AIO term mentions for Papers with Code methods data, grouped by AIO ontology class. The top 20 classes with the most mentions are shown and the remainder are summarized in the final 'Other' category.

The results of this evaluation are summarized by the number of mentions of AIO classes within the Papers with Code methods data (Figure 2). There were a total of 6,484 AIO annotations for the Papers with Code Methods data. 4,647 of these annotations were exact matches to an AIO ontology class label, 1,837 annotations were to exact synonyms.

Within the mentions of AIO terms in the Papers with Code methods data, we observed significant coverage of AIO classes related to deep learning architectures, data preprocessing techniques, and ethical considerations in AI, among others. This NLP evaluation demonstrates AIO's alignment with current AI research trends but also highlights potential areas for ontology refinement and extension. We can use this evaluation as a reproducible process to continuously update AIO in the future.

## Example application: enhanced model cards

Enhancing the Model Cards concept (Mitchell et al., 2019) is an example of how AIO can be leveraged to improve transparency and understanding of AI models. Model Cards, which document the performance characteristics and intended uses of AI models, can be enhanced by the standardized terminology and concepts provided by AIO. By incorporating AIO into Model Cards, developers can offer more detailed and comprehensible descriptions of their models' architectures, functionalities, and ethical considerations. In turn, model users and the public can benefit from these standardized records. This not only facilitates better communication within the research community but also promotes responsible AI development and deployment by ensuring that AI model users have a clear understanding of a model's capabilities and limitations.

# Discussion

## Use cases

The use cases for AIO are diverse, reflecting its broad applicability across AI research and development. From enhancing transparency in AI methodologies to facilitating the annotation and comparison of AI models, AIO has the potential to serve as a foundational tool for adding transparency AI technologies. It can enable researchers to identify sets of methods or publications referring to a standardized AI term, along with more complex applications such as comparing model code implementations and developing formal distance measures between AI/ML methods. By standardizing AI terminology, AIO supports the annotation of code repositories and academic papers. We consider this terminological consistency to be one strategy toward clarifying communication between researchers, particularly as AI methods are adopted in new domains. Providing a standardized methodological vocabulary may also assist AI research newcomers in understanding publications in this often opaque field (Kocak et al., 2021), especially the links between method code and method descriptions and method classifications.

## Alignment with OBO

AIO aligns with Open Biological and Biomedical Ontology (OBO) Foundry (Jackson et al., 2021) standards and principles. AIO is not intended to be part of the OBO Foundry, as its scope extends beyond biology, it does not use OBO IDs, and does not conform to an upper ontology. Terms in AIO may nevertheless be imported by OBO ontologies using standard workflows.

## Ongoing maintenance

Ongoing maintenance of AIO is crucial for its relevance and utility in the fast-paced field of AI. This maintenance involves regular updates to incorporate new AI methodologies, tools, and ethical considerations, ensuring that the ontology accurately reflects current practices and advancements. The process is supported by AI-driven tools that facilitate the identification and integration of emerging concepts.There are a series of strategies in place for keeping AIO up to date. Some of these are technical, like how the build process is reproducible with ODKs. Some are social, like following OBO strategies and the minimum information for reporting an ontology (MIRO) (Matentzoglu et al., 2018), such that AIO integrates well with other resources and should continue to do so. In addition, a number of in progress or near future strategies involve using the Ontology Access Kit (OAK) (Mungall et al., 2022), for example automating lexical mappings and mining literature for new candidate ontology classes, term synonyms, as well as novel AI model architectures.

## Limitations

The limitations of AIO primarily concern its scope and the inherent complexity of AI technologies. AIO aims to cover a broad range of AI concepts but does not delve into the specifics of individual model implementations or parameter values, which can vary widely across different applications. This limitation is intentional, to manage complexity and maintain the ontology's usability. There is also a simplification in AIO regarding modeling Network Layers, as these are represented as a list but may have more sophisticated, nonlinear architectures, e.g. with loops. Another limitation is that while the ontology is designed with composable concepts, this may not yet suffice for supporting full concept composability due to the previously mentioned lack of parameters but also because method composability requires alignment with respect to the input data and for now that aspect is out of scope for AIO. Future developments may address these aspects to further enhance AIO's applicability.

## AI model and AI research publication catalogs

The advent of catalogs for AI publications and models such as Papers with Code, Hugging Face, as well as code repositories such as GitHub, marks a crucial step towards standardization in the rapidly evolving AI field. These resources represent different substrates for annotation with AIO ontology terms, and are becoming key resources for promoting reproducibility, enhancing collaboration, and ensuring that innovations are easily accessible. The incorporation of AIO into these platforms, or for analysis of data from these platforms, could further standardize AI terminology and concepts.

## Conclusions

AIO advances the standardization and understanding of AI concepts and methodologies. By providing a comprehensive framework for AI terminology, AIO facilitates clearer communication, collaboration, and sharing of results within the AI community. Its applications, from enhancing model cards to supporting ongoing projects, demonstrate its value across different domains. AIO's sustainability is supported by regular updates, community contributions, and adherence to best practices in ontology management. As the field of AI continues to expand, the framework supporting the AIO development and maintenance should be better equipped to keep pace, thanks to the automated ontology validation, integration with multiple use cases, and LLM-assisted possibilities for ontology extension and maintenance. AIO can help ensure that AI reports, comparisons, and advancements are grounded in a standardized and accessible vocabulary.

## Acknowledgements

The development of the AIO was supported by funding from the ENDURABLE project DE-AC02-05CH11231.